\documentclass[journal,12pt,onecolumn,draftclsnofoot,]{IEEEtran}
\usepackage{amsmath,amsfonts}
\usepackage{algorithmic}
\usepackage{algorithm}
\usepackage{array}
\usepackage[caption=false,font=normalsize,labelfont=sf,textfont=sf]{subfig}
\usepackage{textcomp}
\usepackage{stfloats}
\usepackage{url}
\usepackage{verbatim}
\usepackage{graphicx}
\usepackage{cite}
\usepackage{bbm}
\begin{document}

\title{Calibration-Aware Bayesian Learning}


\author{\IEEEauthorblockN{Jiayi Huang, Sangwoo Park, and Osvaldo Simeone}
\IEEEauthorblockA{\\King’s Communications, Learning \& Information Processing (KCLIP) Lab \\
}
\thanks{The work of Osvaldo Simeone and Sangwoo Park were supported by the European Research Council (ERC) under the European Union’s Horizon 2020 Research and Innovation Programme (grant agreement No. 725732), by the European Union’s Horizon Europe project CENTRIC (101096379), and by an Open Fellowship of the EPSRC (EP/W024101/1). The work of Jiayi Huang was supported by the China Scholarship Council and King’s College London for their Joint Full-Scholarship (K-CSC) under Grant CSC No. 202206150005. Preprint submitted to IEEE MLSP 2023.}}




\maketitle

\begin{abstract}
Deep learning models, including modern systems like large language models, are well known to offer unreliable estimates of the uncertainty of their decisions. In order to improve the quality of the confidence levels, also known as calibration,  of a model, common approaches entail the addition of either data-dependent or data-independent  regularization terms to the training loss. Data-dependent regularizers have been recently introduced in the context of conventional frequentist learning to penalize deviations between confidence and accuracy. In contrast, data-independent regularizers are at the core of Bayesian learning, enforcing adherence of the variational distribution in the model parameter space to a prior density. The former approach is unable to quantify epistemic uncertainty, while the latter is severely affected by model misspecification. In light of the limitations of both methods, this paper proposes an integrated framework, referred to as calibration-aware Bayesian neural networks (CA-BNNs), that applies both regularizers while optimizing over a variational distribution as in Bayesian learning. Numerical results validate the advantages of the proposed approach in terms of expected calibration error (ECE) and reliability diagrams. 
\end{abstract}
\vspace{-0.2cm}
\section{Introduction}
\label{sec:intro}
\vspace{-0.2cm}
For deep learning tools to be widely adopted in applications with strong reliability requirements, such as  engineering or health care, it is critical that data-driven models   be able to quantify the likelihood of producing incorrect decisions \cite{hullermeier2021aleatoric,openai2023}. This is currently an open challenge for  conventional \emph{frequentist learning}, which  is known to produce overconfident, and hence poorly calibrated, outputs, especially in the presence of limited training data \cite{guo2017calibration}. This paper contributes to the ongoing line of work concerned with the introduction of novel methodologies for the design of well-calibrated machine learning models.

The gold standard of calibrated machine learning is set, under ideal conditions, by 
\emph{Bayesian learning}, which treats the model parameters as random variables. In Bayesian learning, the distribution over the model parameters is optimized by introducing a \emph{data-independent}, information-theoretic,  regularizer that enforces adherence to a prior distribution (see, e.g., \cite{simeone2022machine}).  The optimized distribution is then used to make decisions via \emph{ensembles} of models that account for the epistemic uncertainty caused by the limited availability of data. However, when the model -- prior distribution and likelihood function -- are \emph{misspecified}, Bayesian learning is no longer guaranteed to provide well-calibrated decisions  \cite{knoblauch2019generalized, masegosa2020learning, wenzel2020good}.  In practice, model misspecification is hard to ascertain, and hence it is important to develop versions of Bayesian learning that more directly address the criterion of calibration.

In a separate line of work, recent studies \cite{kumar2018trainable, bohdal2021meta} have shown that introducing a \emph{data-dependent} regularizer that penalizes calibration errors can improve the calibration performance of conventional frequentist learning. However, these studies are limited to decisions made using single models, and they are thus by design not suitable to capture \emph{epistemic uncertainty} by means of ensembling over multiple models as in Bayesian learning.

In light of the mentioned limitations of both approaches, this paper proposes an integrated training framework, referred to as \emph{calibration-aware Bayesian neural networks} (CA-BNNs). As described in Sec. 4, after providing the necessary background in Sec. 2 and Sec. 3, the proposed training criterion  applies  a data-dependent regularizer that penalizes calibration errors, as in \cite{kumar2018trainable, bohdal2021meta}, as well as a data-independent regularizer enforcing adherence to a prior density, while optimizing over a variational distribution, as in Bayesian learning. As a secondary contribution, in Sec. 5, we also describe an improvement to the training strategy introduced in \cite{kumar2018trainable} that relies on fully differentiable calibration error metrics \cite{bohdal2021meta, park2022few}.  Experiments presented in Sec. 6 validate the proposed approach.

\section{BACKGROUND}
\label{sec:background}

As introduced in Sec.~\ref{sec:intro}, this paper proposes a variant of Bayesian learning for neural networks that directly addresses the performance criterion of calibration. The proposed  CA-BNN training framework builds on a frequentist counterpart, introduced in \cite{kumar2018trainable}, which we will refer to as  \emph{calibration aware-frequentist neural networks} (CA-FNNs). As mentioned in Sec.~\ref{sec:intro} and reviewed in Sec.~\ref{sec:ca-fnn}, CA-FNN adds a \emph{batch regularizer} to the standard cross-entropy training loss minimized by frequentist learning that provides an estimate of the calibration error. In order to provide the necessary background, this section reviews frequentist and Bayesian learning (see, e.g., \cite{simeone2022machine}), as well as the standard measure of calibration known as the \emph{expected calibration error} (ECE) \cite{guo2017calibration}.
\subsection{Frequentist and Bayesian learning}
We study a conventional supervised learning formulation in which the goal is to predict a discrete output variable $y\in\mathcal{Y}$ given an input variable $x\in\mathcal{X}$. We fix a parameterized class $p(y|x,\theta)$ of classifiers, e.g., a neural network with a softmax output layer. Training and testing data follow an (unknown) joint distribution $p(x,y)$. We are given a training data set $\mathcal{D} = \{{(x_{i}, y_{i})}\}^n_{i = 1}$, with $i$-th input $x_i \in \mathcal{X}$ and corresponding output $y_i \in \mathcal{Y}$.

\subsubsection{Frequentist Learning}

Based on the training data, conventional \emph{frequentist learning} finds a single parameter vector $\theta$, and consequently a single classifier  $p(y|x,\theta)$, for use on the test data. This is done by  minimizing the training loss $\mathcal{L}(\theta|\mathcal{D})$ as per the problem \begin{align}
    \label{eq:freq_obj}
     \theta^\text{tr} = \arg\min _{\theta} \mathcal{L}(\theta|\mathcal{D}), 
\end{align} where the training loss is typically given by the \emph{cross entropy} \begin{align}
    \label{eq:ce_loss}
    \mathcal{L}(\theta|\mathcal{D}) = - \sum_{(x_i,y_i)\in\mathcal{D} }\log p(y_i|x_i,\theta). 
\end{align}

Frequentist learning is known to yield trained probabilistic predictors $p(y|x,\theta^\text{tr})$ that are poorly calibrated, especially in the presence of limited training data \cite{guo2017calibration}. This indicates, as we will formalize later in this section, that the probability $p(y|x,\theta^\text{tr})$ assigned to an output value $y$ given the input $x$ is not a good estimate of the true probability that the output variable equals $y$ given input $x$.


\subsubsection{Bayesian Learning}
\label{subsec:bayesian}
The poor calibration of frequentist learning may be ascribed to its  reliance of a single model parameter $\theta^\text{tr}$. In fact, this choice may result in discarding uncertainty in the model space that arises due to the limited availability of data. Bayesian learning captures such \emph{epistemic uncertainty} in the model parameter space by treating the model parameter vector $\theta$ as a \emph{random} vector. Specifically, taking a \emph{variational inference} (VI) perspective on Bayesian learning, the model parameter vector is assigned a  parameterized distribution $q(\theta|\varphi)$ with parameter vector $\varphi$ \cite{simeone2022machine, angelino2016patterns}. As we detail next, in VI, the  parameter vector $\varphi$ is optimized, obtaining vector $\varphi^\text{tr}$,  based on the training data $\mathcal{D}$.  

By maintaining a distribution $q(\theta|\varphi)$ on the model parameter vector $\theta$, Bayesian learning allows decisions to be made not based on a single classifier, but rather via an \emph{ensemble} of classifiers sampled from distribution $q(\theta|\varphi^\text{tr})$ as per the expectation\begin{align}
    \label{eq:bayesian_predictive}
   p(y |x,\mathcal{D}) = \mathbb{E}_{\theta \sim q(\theta|\varphi^\text{tr})} [p(y|x,\theta)].
\end{align}
The use of ensembling allows prediction \eqref{eq:bayesian_predictive} to better account for epistemic uncertainty, typically improving, as a result, the calibration of the classifier \cite{blundell2015weight, ravi2019amortized}.


Let us fix a prior distribution $p(\theta)$ on the model parameters \cite{blundell2015weight,wenzel2020good}. Following VI, the parameter vector  $\varphi^\text{tr}$ is obtained  by minimizing the \emph{free energy} (see, e.g., \cite{simeone2022machine}) as per the problem \begin{align} \label{bnn}
    \min_{\varphi} \Big\{ \mathbb{E}_{\theta \sim q(\theta |\varphi)} [\mathcal{L}(\theta|\mathcal{D})]  + \beta \cdot \operatorname{KL}(q(\theta |\varphi) \| p(\theta)) \Big\},
\end{align}
where $\operatorname{KL} (q(\theta) \| p(\theta)) = \mathbb{E}_{q(\theta)} [\log (q(\theta)/p(\theta))]$ is the Kullback-Liebler (KL)  divergence and $\beta \geq 0$ is a hyperparameter. Note that the KL term is a \emph{data-independent regularizer} that mitigates overfitting and poor calibration by enforcing adherence to a prior distribution. 

To address problem (\ref{bnn}), one often chooses the variational distribution $q(\theta|\varphi)$ as a Gaussian distribution with mean and covariance defining the model parameters $\varphi$.  This way, one can apply gradient descent via the reparametrization trick as we detail in Sec.~\ref{sec:calibration_bayesian} \cite{blundell2015weight}.




\vspace{-0.3cm}
\subsection{Expected Calibration Error}
\label{section_ece}
\vspace{-0.12cm}
Consider a trained classifier $p^\mathrm{tr}(y |x)$, which may be $p(y |x,\theta^\mathrm{tr})$ for frequentist learning or $p(y |x,\mathcal{D})$ in (\ref{eq:bayesian_predictive}) for Bayesian learning. Given an input $x$, a \emph{point classification decision} $\hat{y}$ is typically determined by the classifier as the value of $y$ that maximizes the model's confidence, i.e.,
\begin{align} \label{yhati}
    \hat{y}(x) = \arg \max_{y \in \mathcal{Y}} p^\mathrm{tr}(y |x).
\end{align}
Accordingly, the classifier assigns \emph{confidence level} $p^\mathrm{tr}(\hat{y}(x)|x)$ to such decision.  

A \emph{perfectly calibrated classifier} satisfies the condition \cite{guo2017calibration}
\begin{align} \label{calibrated}
    \Pr (y=\hat{y}(x) |p^\text{tr}(\hat{y}(x) |x) = \pi) = \pi, \text{ for all }  \pi \in [0,1],
\end{align}
in which the probability is taken over the true (unknown) joint distribution $p(x,y)$ of the input-output pair $(x,y)$. Condition \eqref{calibrated} says that, on average, for all inputs $x$ with decision (\ref{yhati}) having confidence $\pi$, the true probability the decision being correct is indeed $\pi$. In words, the \emph{confidence level} $\pi$ of the classifier matches the \emph{true accuracy} of the decision.


In order to evaluate the extent to which a classifier satisfies the condition \eqref{calibrated}, reference \cite{guo2017calibration} introduced the ECE. To describe it, divide the confidence interval $[0,1]$ into $M$ intervals $B_m = [(m-1)/M, m/M]$ for $m = 1,...,M$. Then, assign to the $m$-th  data bin, $B_m$, the indices $i$ of the input samples $x_i$ whose \emph{confidence score}
\begin{equation} \label{confid_score} r_{i} = p^\mathrm{tr}(\hat{y}(x_i)|x_i)\end{equation}
falls within the $m$-th interval $B_m$. For each input $x_i$, define also the \emph{correctness score}  
\begin{equation} \label{correct_score} c_{i} = \mathbbm{1} (\hat{y}(x_i) =  y_{i}), \end{equation} 
given the indicator function $\mathbbm{1} (\cdot)$ with $\mathbbm{1} (\text{true}) = 1$ and $\mathbbm{1} (\text{false}) = 0$. 

For each bin $m$, the ECE computes the \emph{per-bin confidence} as $\mathrm{conf}(B_m) ={1}/{|B_{m}|} \sum_{i \in B_m} r_i$, and the \emph{per-bin accuracy} as $\mathrm{acc}(B_m) = {1}/{|B_{m}|} \sum_{i \in B_m}  c_i$. \emph{Reliability diagrams} plot the accuracy $\mathrm{acc}(B_m)$ and confidence level $\mathrm{conf}(B_m)$ as a function of the bin number (see Sec.~\ref{sec:results}) \cite{guo2017calibration}. The ECE provides a scalar measure of calibration by measuring the weighted sum of the differences between per-bin confidence and accuracy levels as
\begin{align} \label{ece}
    \mathrm{ECE} = \sum^{M}_{m=1} \frac{|B_{m}|}{ \sum_{m'=1}^M |B_{m'}|} \left | \mathrm{acc}(B_{m}) - \mathrm{conf}(B_m) \right |.
\end{align}

\section{CALIBRATION-AWARE FREQUENTIST LEARNING}
\label{sec:ca-fnn}

Reference \cite{kumar2018trainable} introduces a calibration-aware frequentist learning approach, referred here as CA-FNN, that is based on regularizing the cross entropy \eqref{eq:ce_loss} with a data-dependent regularizer that provides an estimate of the ECE described in Sec.~\ref{section_ece}. In this section, we provide a brief description of CA-FNN for reference.

Let $\mathrm{AECE}(\theta|\mathcal{D})$ denotes a \emph{differentiable} approximation of the ECE (\ref{ece}) obtained by replacing the trained model $p^\mathrm{tr}(y|x)$ in \eqref{yhati} with a classifier $p(y|x,\theta)$ with an arbitrary model parameter $\theta$. The notation $\mathrm{AECE}(\theta|\mathcal{D})$ makes the dependence on the data set $\mathcal{D}$ and on the parameter $\theta$ explicit. Note that the ECE \eqref{ece} is not differentiable, and hence it cannot be directly used as an optimization criterion in standard gradient-based optimizers, requiring the introduction of the approximation $\mathrm{AECE}(\theta|\mathcal{D})$. Reference \cite{kumar2018trainable} focused specifically on an estimate, referred to as WMMCE, which will be described in Sec.~\ref{subsec:WMMCE}.


CA-FNN, as introduced in \cite{kumar2018trainable}, addresses the  problem 
\begin{align} \label{ca-fnn}
 \min _{\theta}  \mathcal{L}(\theta|\mathcal{D})+ \lambda \cdot  \textrm{AECE} (\theta|\mathcal{D}) 
 \end{align} 
 for some hyperparameter $\lambda>0$. The rationale for the approach is that problem (\ref{ca-fnn}) not only aims at minimizing the training loss, but also a measure of the calibration error. 

 \section{CALIBRATION-AWARE BAYESIAN LEARNING}
\label{sec:calibration_bayesian}

As discussed in Sec.~\ref{subsec:bayesian}, by estimating epistemic uncertainty via ensembling, Bayesian learning can generate better calibrated models as compared to frequentist learning. However, it is well known that the improvements in calibration brought by Bayesian learning are predicated on the assumption that the model --  prior distribution and  likelihood function -- are well specified, providing a sufficiently accurate match with the ground-truth data generation distribution  \cite{knoblauch2019generalized, wenzel2020good, zecchin2022robust}. In light of this limitation, we propose to integrate Bayesian learning with ECE-based regularization, in a manner akin to CA-FNN, in order to enhance the calibration of neural networks trained via Bayesian learning. Accordingly, we refer to the proposed approach as CA-BNN. 

CA-BNN is based on the optimization of a \emph{calibration-aware free energy}, which augments the free energy \eqref{bnn} minimized by VI-based Bayesian learning  with a differentiable estimate of the ECE. Accordingly, we propose to minimize the calibration-aware free energy $\mathcal{F}^\text{CA}(\varphi|\mathcal{D})$, which is defined as 
\begin{align} \label{ca-bnn}
\min_{\varphi} \Big\{ \mathcal{F}^{CA}(\varphi | \mathcal{D}) = &\mathbb{E}_{\theta \sim q(\theta |\varphi)}    [\mathcal{L}(\theta|\mathcal{D}) +  \lambda \cdot \textrm{AECE} (\theta|\mathcal{D})]  + \beta \cdot \operatorname{KL}(q(\theta |\varphi) \| p(\theta)) \Big\},
\end{align} 
where $\textrm{AECE} (\theta|\mathcal{D})$, as defined in the previous section, is a differentiable approximation of the ECE for the model $p(y|x,\theta)$ (see next sections for an example).




We set the distribution under optimization as the Gaussian density \cite{blundell2015weight}
\begin{align}
    \label{eq:vi_mean_field}
    q(\theta|\varphi) = \mathcal{N}(\theta|\mu, \text{Diag}(\exp{(2 \rho)})),
\end{align}
with $N\times 1$ mean vector $\mu$ and $N \times 1$ log-standard deviation vector $\rho$. Hence, the variational parameter $\varphi$ is defined as $\varphi = [\mu, \rho]$. By \eqref{eq:vi_mean_field}, the random vector $\theta$ can be expressed as $\theta = \mu + \exp{(\rho)}\circ \epsilon$, where $\circ$ is the element-wise product, and we have the standard Gaussian random vector $\epsilon \sim \mathcal{N}(0_N, I_N)$, with all-zero mean vector $0_N$ and covariance equal to the identity matrix $I_N$.


    Using this expression, by the reparametrization trick \cite{blundell2015weight}, the expectation term in \eqref{ca-bnn} can be replaced with expectation over standard Gaussian distribution. Accordingly, gradient-based optimization for problem \eqref{ca-bnn} can be realized by producing $R$ independent samples $\theta_{r} = \mu + \exp{(\rho)}\circ \epsilon_r$ with independent vectors $\epsilon_r \sim \mathcal{N}(0_N, I_N)$ for $r=1,...,R$, and by approximating the gradient $\nabla_{\varphi}\mathcal{F}^{CA}(\varphi | {\mathcal{D}})$ using a mini-batch data set $\tilde{\mathcal{D}} \subseteq \mathcal{D}$ as 
\begin{align}
    \label{eq:grad_ca_bnn}
    &\nabla_{\varphi}\mathcal{F}^{CA}(\varphi | {\Tilde{\mathcal{D}}}) \approx \frac{1}{R} \sum^{R}_{r=1} \Big[ \left(\nabla_{\varphi}\theta_r\right) \nabla_{\theta_r}\big(\mathcal{L}(\theta_r|\Tilde{\mathcal{D}}) + \lambda \cdot \textrm{AECE} (\theta_r|\Tilde{\mathcal{D}})\big)\Big] + \beta \frac{ |\tilde{\mathcal{D}}| }{|\mathcal{D}|} \cdot \nabla_{\varphi} \operatorname{KL}(q(\theta |\varphi) \| p(\theta)).
\end{align}
Note than in \eqref{eq:grad_ca_bnn} the calibration-driven regularizer $\textrm{AECE} (\theta|\tilde{\mathcal{D}})$ is evaluated using the mini-batch $\Tilde{\mathcal{D}}$ rather than the entire data set $\mathcal{D}$, and that the scaling by $|\tilde{\mathcal{D}}|/|\mathcal{D}|$ ensures an unbiased estimate of the gradient (see, e.g., \cite[Sec.~3.4]{blundell2015weight}).  

Finally, the variational parameter $\varphi$ is updated as
\begin{align}
    \varphi \leftarrow \varphi - \gamma \nabla_{\varphi} \mathcal{F}^{CA}(\varphi | {\Tilde{\mathcal{D}}}),
\end{align}
with step size $\gamma > 0$ using the gradient estimate \eqref{eq:grad_ca_bnn}. The optimization procedure for CA-BNN is summarized in Algorithm~\ref{alg:cabnn}.

\vspace{-0.3cm}
\begin{figure}[htb]
  \renewcommand{\algorithmicrequire}{\textbf{Input:}}
  \renewcommand{\algorithmicensure}{\textbf{Output:}}
  \begin{algorithm}[H]
    \caption{CA-BNN Training Procedure}
    \begin{algorithmic}[1]
      \REQUIRE Training data set $\mathcal{D}$, ensembling size $R$, prior distribution $p(\theta)$, step size $\gamma$, free energy hyperparameter $\beta$, calibration hyperparameter $\lambda$
      \ENSURE distribution $q(\theta|\varphi)$ over the variational parameter $\varphi = [\mu, \rho]$
      \WHILE{not done}
      \STATE Sample mini-batch $\tilde{\mathcal{D}}$ from $\mathcal{D}$ 
      \FOR{$r= 1,...,R$}
      \STATE Draw $\epsilon_r \sim \mathcal{N}(0_N, I_N)$
      \STATE Reparameterize $\theta_{r} \leftarrow \mu + \exp{(\rho)}\circ \epsilon_r$
      \STATE Compute stochastic gradient \\$g_r = \left(\nabla_{\varphi}\theta_r \right)\nabla_{\theta_r}[\mathcal{L}(\theta_r|\Tilde{\mathcal{D}}) +  \lambda \cdot \textrm{AECE} (\theta_r|\Tilde{\mathcal{D}})]$
      \ENDFOR   
      \STATE Update \\
          $\varphi \leftarrow \varphi - \gamma \left (\frac{1}{R} \sum^{R}_{r=1} g_r + \beta\frac{|\tilde{\mathcal{D}}|}{|\mathcal{D}|} \cdot \nabla_{\varphi} \operatorname{KL}(q(\theta |\varphi) \| p(\theta)) \right )$
      \ENDWHILE
      \STATE {\textbf{return } $q(\theta|\varphi)$}
    \end{algorithmic}
    \label{alg:cabnn}
  \end{algorithm}
  \vspace{-1cm}
\end{figure}

\section{DIFFERENTIABLE CALIBRATION MEASURES}
In this section, we review and extend the differentiable approximate ECE measure, $\textrm{AECE} (\theta|\mathcal{D})$, introduced in \cite{kumar2018trainable} with the name \emph{weighted MMCE} (WMMCE). As discussed in the previous two sections, this measure can be used in the learning objectives \eqref{ca-fnn} and \eqref{ca-bnn} for CA-FNN and CA-BNN, respectively. After reviewing the WMMCE score in \cite{kumar2018trainable}, we introduce an extension that will be shown in the next section 
to be potentially beneficial in improving the calibration performance. We finally note that an alternative differentiable metric was introduced in \cite{bohdal2021meta}, which was found in our experiments (not reported here) to offer similar performance as the WMMCE when used for CA-FNN and CA-BNN.



\subsection{WMMCE}
\label{subsec:WMMCE}

In \cite{kumar2018trainable}, the WMMCE metric was defined as an estimate of the ECE \eqref{ece}. To introduce it, define as $\kappa(\cdot,\cdot)$ a kernel function operating on scalar inputs, such as $\kappa(a,b) = \text{exp}(-|a-b|/\gamma)$ for some $\gamma>0$. Using the training data set $\mathcal{D}$, the WMMCE first computes confidence scores $\{r_i\}_{i=1}^n$ and correctness scores $\{c_i\}_{i=1}^n$ as defined in \eqref{confid_score} and \eqref{correct_score}, respectively, with the parametric classifier $p(y|x,\theta)$ in lieu of the trained classifier $p^\text{tr}(y|x)$ as discussed in Sec.~\ref{sec:ca-fnn}. Then, the WMMCE evaluates the metric
\begin{align} \label{weighted}
 \mathrm{AECE} (\theta | \mathcal{D})   = \Bigg(\sum_{i,j:{c}_i = {c}_j = 0} \frac{{r}_i {r}_j \kappa({r}_i , {r}_j)}{(n-n_c)(n-n_c)} & + \sum_{i,j:{c}_i = {c}_j = 1} \frac{(1-{r}_i) (1-{r}_j) \kappa({r}_i , {r}_j)}{n_c ^2} \nonumber \\
 & -2 \sum_{i,j:{c}_i = 1,  {c}_j = 0} \frac{(1-{r}_i) {r}_j \kappa({r}_i , {r}_j)}{(n-n_c)n_c} \Bigg)^{\frac{1}{2}},
\end{align}
where $n_c = \sum_{i = 1}^n {{c}}_{i}$ is the number of correct examples. The sums in \eqref{weighted} are extended over all examples in data set $\mathcal{D}$.




\subsection{Gradient of the WMMCE}
\label{differentiable_section}
In order to evaluate the gradient $\nabla_{\theta}\textrm{AECE} (\theta|\tilde{\mathcal{D}})$ required for both CA-FNN and CA-BNN, one needs to calculate the gradients $\nabla_{\theta} r_i$ and $\nabla_{\theta} c_i$ of the confidence and correctness scores, respectively. To simplify this calculation, reference \cite{kumar2018trainable} implicitly ignored the dependence of the point classification decision $\hat{y}(x)$ in \eqref{yhati} on the model parameter $\theta$. Accordingly,  the gradient of the correctness score was set to zero; and  the gradient $\nabla_{\theta} r_i$ was evaluated as $\nabla_{\theta} p (\hat{y}(x) | x, \theta)$, where $\hat{y}(x)$ is treated as a constant. This approximation of the gradient is motivated by the non-differentiable nature of the decision $\hat{y}(x) = \arg \max_{y \in \mathcal{Y}} p(y | x, \theta)$ with respect to $\theta$. 

\begin{figure}[tb]
    \centering
    \centerline{\includegraphics[scale=0.6]{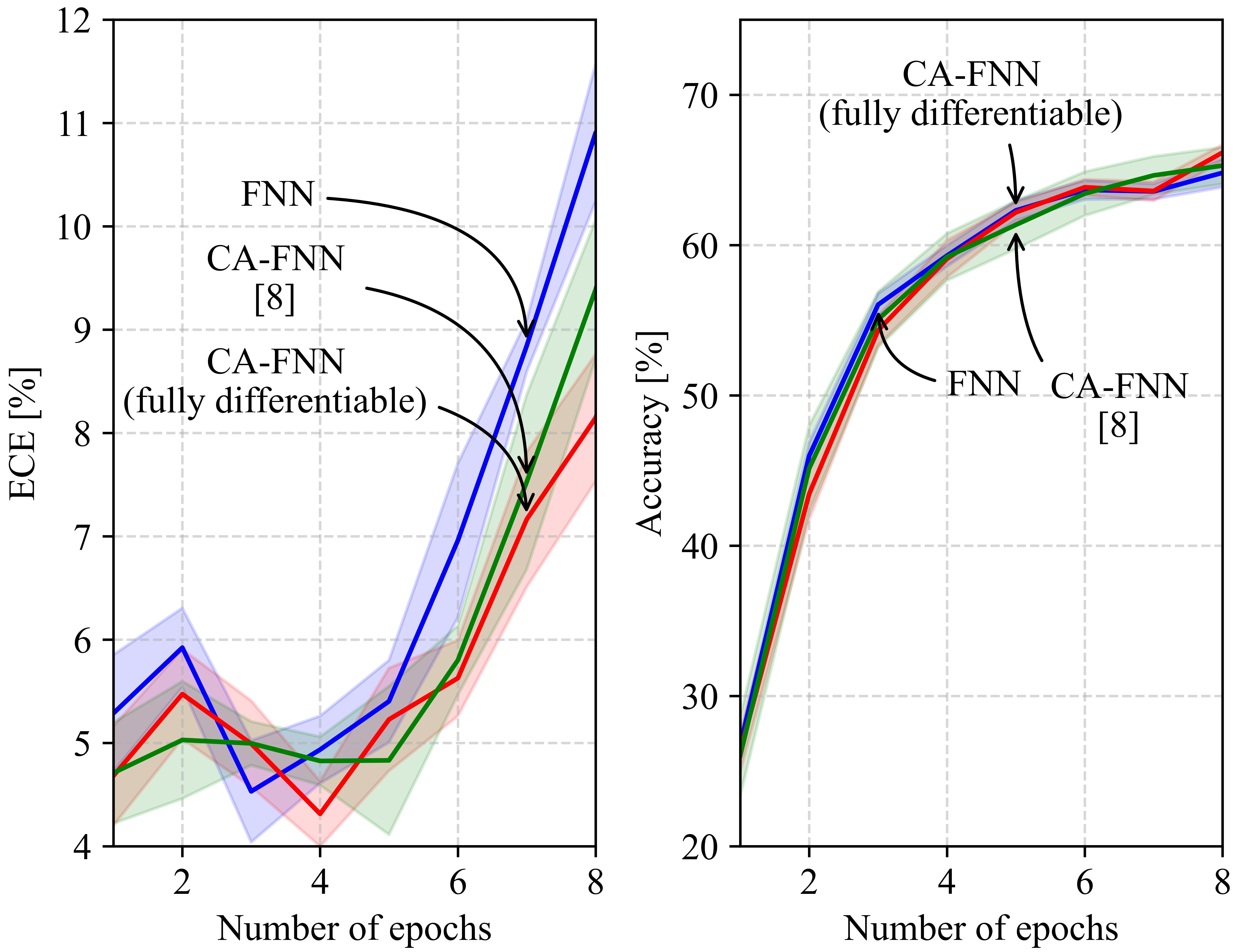}}
    \caption{ECE and accuracy as a function of number of epochs for 20 Newsgroups classification task for FNN, CA-FNN \cite{kumar2018trainable}, and the modified CA-FNN introduced in Sec.~\ref{differentiable_section} with fully differentiable batch regularizer. The shaded areas correspond to  $75\%$  intervals of the realized values.}
    \label{fig:1}
\end{figure}
In this section, as a secondary contribution, we propose potentially more accurate estimators of the gradient of the confidence and correctness scores, replacing the non-differentiable maximum operator in \eqref{yhati} with differentiable \emph{smoothed maximum} operator \cite{park2022few, cuturi2019differentiable}.

Specifically, the \emph{differentiable confidence score} is defined by taking the smoothed maximum among all the available candidate confidence scores $\{p (y |x_i,\theta)\}_{y \in \mathcal{Y}}$ as
\begin{align} \label{rhat}
    \hat{r_i} = \sum_{y \in \mathcal{Y}} p (y |x_i,\theta) \frac{e^{p (y |x_i,\theta) / \tau_r}}{\sum_{y' \in \mathcal{Y}} e^{p (y' |x_i,\theta) / \tau_r}},
\end{align}
with temperature parameter $\tau_r > 0$ controlling the smoothness of the approximation (\ref{rhat}). In the limit $\tau_r \rightarrow 0$, the differentiable confidence score $\hat{r_i}$ recovers the true score $r_i$. Furthermore, we adopt the \emph{differentiable correctness score} \cite{bohdal2021meta}
\begin{align}
    \label{chat}
    \hat{c_i} = \text{ReLU} (2 - [\hat{R}(x_i)]_{y_i})
\end{align}
with $\text{ReLU}(a)=\max(0,a)$ and  differentiable rank function
\begin{align} \label{Rhat}
    [\hat{R}(x)]_{y} = 1 + \sum_{y^{\prime} \in \mathcal{Y}, y \neq y^{\prime}} \frac{1}{1 + e^{S_{y, y^{\prime}} / \tau_c}},
\end{align}
with $S_{y, y^{\prime}} = p(y |x, \theta) - p(y^{\prime} |x, \theta)$. With small enough temperature parameter $\tau_c > 0$, the differentiable correctness score $\hat{c_i}$ recovers the true correctness score $c_i$ in \eqref{correct_score}. 

By replacing $r_i, r_j$ and $c_i, c_j$ in \eqref{weighted} with $\hat{r}_i, \hat{r}_j$ and $\hat{c}_i$, $\hat{c}_j$ via \eqref{rhat} and \eqref{chat}, respectively for $i,j=1,...,n$, one obtains a \emph{fully differentiable} version of the AECE, which can be differentiated with respect to the model parameter $\theta$.

\section{EXPERIMENTS AND DISCUSSION}
\label{sec:results}

\begin{figure} [tb] 
    \centering
    \centerline{\includegraphics[scale=0.6]{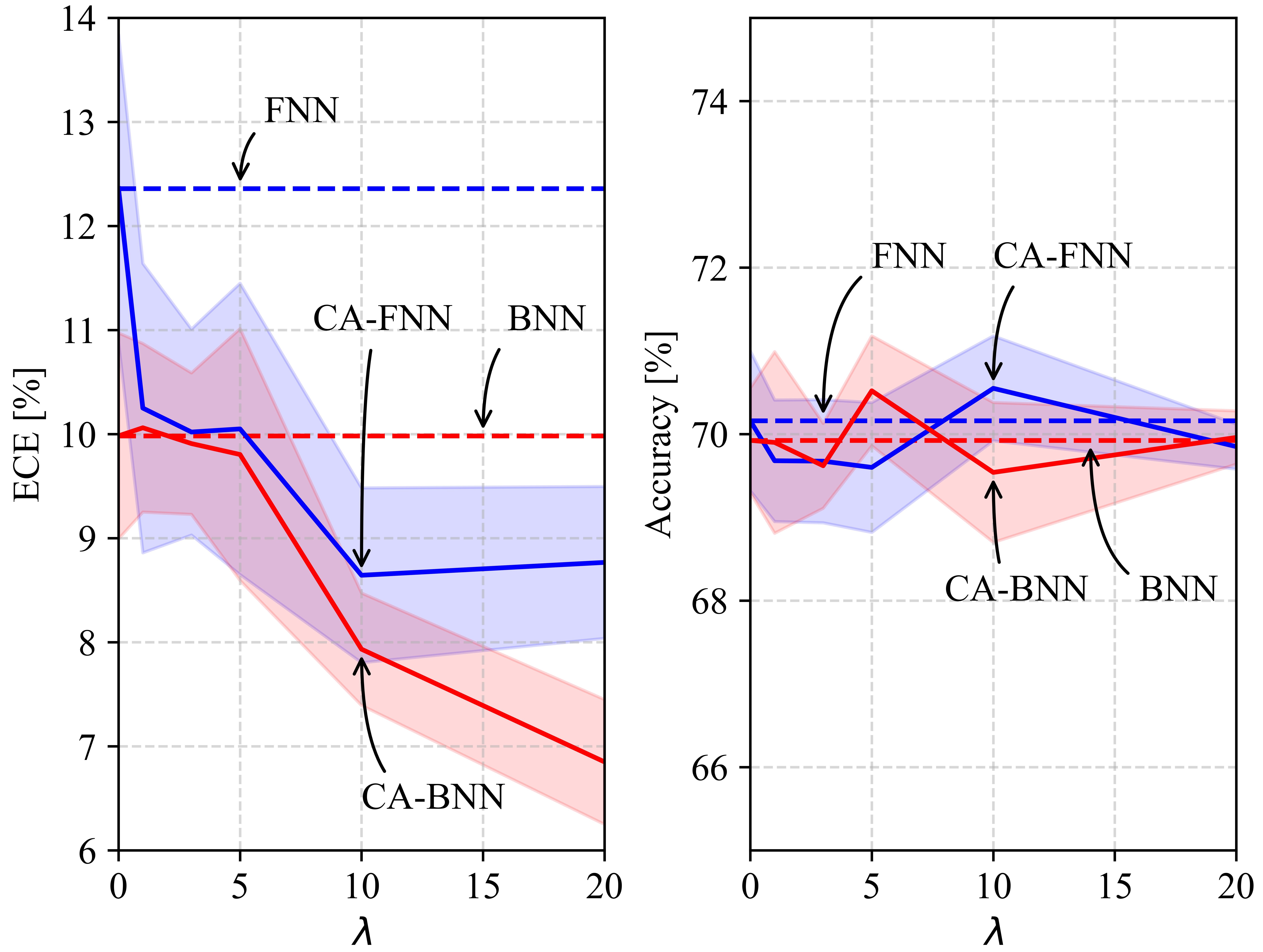}}
    \caption{ECE and accuracy as a function of hyperparameter $\lambda$ on 20 Newsgroups data set for FNN, BNN, CA-FNN and CA-BNN both with fully differentiable batch regularizer. The shaded areas correspond to $60\%$ intervals of the realized values.}
    \label{fig:2} 
\end{figure}

In this section, we compare the calibration performance and the accuracy of conventional FNN and BNN, as well as CA-FNN \cite{kumar2018trainable} and CA-BNN (this work). We validate the proposed framework by considering two popular classification data sets, namely 20 Newsgroups \cite{lang1995newsweeder} and CIFAR-10 \cite{krizhevsky2010cifar}.
\footnote{Code can be found at \url{https://github.com/kclip/CA-BNN}.}.

\subsection{20 Newsgroups}
For the 20 Newsgroups classification task, as in \cite{lin2013network}, we adopt a convolutional neural network with global pooling for all schemes. For Bayesian learning, i.e., for BNN and CA-BNN, we choose $\beta=0.1$ in \eqref{ca-bnn}, with zero-mean Gaussian prior $p(\theta)$ with standard deviation 0.05. The temperature parameters in \eqref{rhat} and \eqref{Rhat} are set to $\tau_r = 0.001$ and $\tau_c = 0.01$; and we set $\gamma=0.4$ for the WMMCE kernel. We use the RMSprop optimizer with learning rate 0.002.


We first examine the impact of the differentiable confidence and correctness scores introduced in Sec.~\ref{differentiable_section}. Fig.~\ref{fig:1} shows ECE and accuracy as a function of number of training epochs for (\emph{i}) FNN; (\emph{ii}) CA-FNN with the batch regularizer proposed in \cite{kumar2018trainable}; and (\emph{iii}) CA-FNN with the fully differentiable batch regularizer introduced in Sec.~\ref{subsec:WMMCE}. We fix the hyperparameter $\lambda$ to 10. It is observed that the calibration performance in terms of ECE is enhanced by the proposed fully differentiable regularizer, while not affecting the accuracy. Therefore, in the following experiments, we only consider fully differentiable confidence and correctness scores.



In Fig.~\ref{fig:2}, we investigate the impact of the hyperparameter $\lambda$ that dictates the trade-off between accuracy and confidence in the training objectives \eqref{ca-fnn} and \eqref{ca-bnn}. Note that setting $\lambda$ to 0 recovers standard FNN and BNN (dashed lines), with the BNN having a lower ECE than FNN \cite{blundell2015weight, zecchin2022robust}. Increasing the value of $\lambda$ has a positive effect on the calibration of both CA-FNN, as also reported in \cite{kumar2018trainable}, as well as of the proposed CA-BNN, while only marginally affecting the accuracy in the range of considered values of $\lambda$. Furthermore, CA-BNN can decrease the ECE as compared to CA-FNN by more than 2\%.

\begin{figure*}[tb]
    \centering
   \centerline{\includegraphics[scale=0.41]{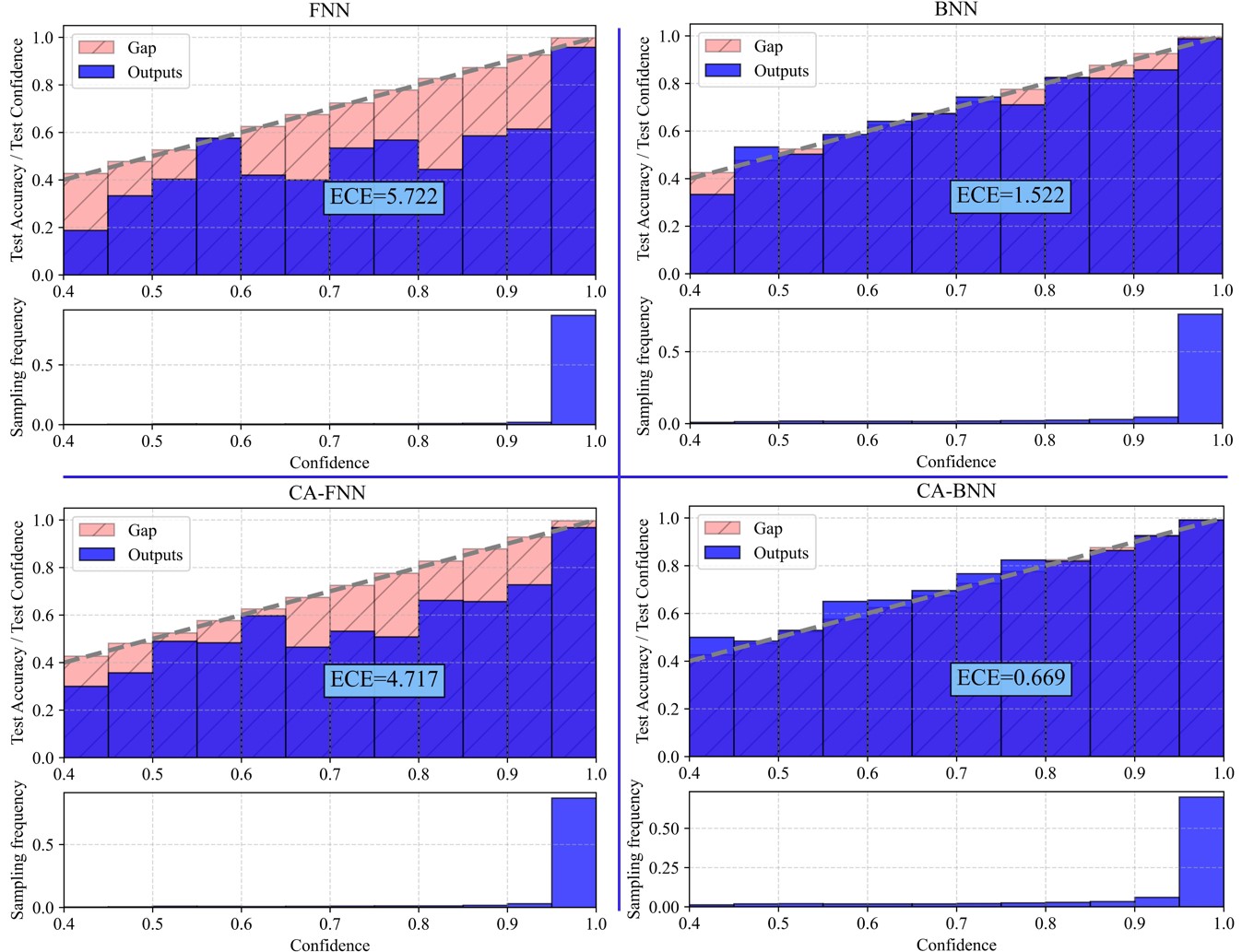}}
   \caption{Reliability diagrams for the CIFAR-10 classification task given the predictor trained by (\emph{i}) FNN (top-left), (\emph{ii}) BNN (top-right), (\emph{iii}) CA-FNN (bottom-left), (\emph{iv}) CA-BNN (bottom-right).}
\label{fig:3}
\end{figure*}
\subsection{CIFAR-10}
For CIFAR-10 classification, we adopt a pre-trained ResNet18 model \cite{he2016deep}, which is fine-tuned via the Adam optimizer with learning rate 0.00001 for three epochs with the entire training data set for all schemes. For Bayesian learning we set $\beta=0.00001$ with zero-mean Gaussian prior with standard deviation $0.001$. Temperature parameters for differentiable measures are set to $\tau_s = 0.001$ and $\tau_c=0.01$; and we set $\gamma=0.4$ for the WMMCE kernel. The hyperparameter $\lambda$ is chosen for CA-FNN and CA-BNN as 20 and 18, respectively, via a non-exhaustive numerical search using the training set.

Fig.~\ref{fig:3} shows the \emph{reliability diagrams} (see Sec.~\ref{section_ece}) for all four schemes. Bayesian learning is observed to yield better calibrated decisions than frequentist learning,  as also manifest in the lower value of the ECE. Calibration-aware training improves the match between accuracy and confidence, resulting in a lower ECE, for both frequentist and Bayesian learning, with CA-BNN achieving the lowest ECE. 

\section{Conclusion}
Overall, while Bayesian learning is provably well calibrated under ideal assumptions on model specification and availability of computational power, the regularization applied by the proposed calibration-aware Bayesian training is observed to offer performance benefits in terms of ECE in practical scenarios characterized by model misspecification and approximate free energy optimization. Future work will investigate performance under distributional shift, as well as robustness to membership attacks \cite{chen2023overconfidence}.

\vspace{-0.3cm}
\bibliographystyle{IEEEtran}
\bibliography{refs}

\end{document}